\documentclass[twocolumn,10pt]{asme2ej}

\usepackage{graphicx} 
\usepackage{hyperref}   
\hypersetup{
	colorlinks=true,
	linkcolor=blue,
	citecolor=blue,
	urlcolor=blue,
}
\usepackage[square,numbers]{natbib}

\usepackage{amsmath}
\usepackage{graphicx}
\usepackage{multirow}
\usepackage{multicol}
\usepackage{bm}
\usepackage{array}
\usepackage{color,soul}
\usepackage[inline]{enumitem}

\title{
Spherical Rolling Robots Design, Modeling, and Control: A Systematic Literature Review}

\author{Aminata Diouf\thanks{Corresponding author}\\
    \affiliation{
	Lab. INIT Robots\\
	Department of Mechanical Engineering\\
	École de technologie supérieure\\
	Montréal, Canada\\
    Email: aminata.diouf.1@ens.etsmtl.ca
    }	
}

\author{Bruno Belzile \\
    \affiliation{
	Lab. INIT Robots\\
	Department of Mechanical Engineering\\
	École de technologie supérieure\\
	Montréal, Canada\\
    Email: bruno.belzile@etsmtl.ca
    }
}

\author{Maarouf Saad \\
    \affiliation{
	Department of Electrical Engineering\\
	École de technologie supérieure\\
	Montréal, Canada\\
    Email: maarouf.saad@etsmtl.ca
    }
}

\author{David St-Onge \\
    \affiliation{
	Lab. INIT Robots\\
	Department of Mechanical Engineering\\
	École de technologie supérieure\\
	Montréal, Canada\\
    Email: david.st-onge@etsmtl.ca
    }
}

\begin{document}
\maketitle    
\begin{abstract}

{\it Spherical robots have garnered increasing interest for their applications in exploration, tunnel inspection, and extraterrestrial missions. Diverse designs have emerged, including barycentric configurations, pendulum-based mechanisms, etc. In addition, a wide spectrum of control strategies has been proposed, ranging from traditional PID approaches to cutting-edge neural networks. Our systematic review aims to comprehensively identify and categorize locomotion systems and control schemes employed by spherical robots, spanning the years 1996 to 2023. A meticulous search across five databases yielded a dataset of 3189 records. As a result of our exhaustive analysis, we identified a collection of novel designs and control strategies. Leveraging the insights garnered, we provide valuable recommendations for optimizing the design and control aspects of spherical robots, supporting both novel design endeavors and the advancement of field deployments. Furthermore, we illuminate key research directions that hold the potential to unlock the full capabilities of spherical robots.}
\end{abstract}

\section{Introduction}\label{s:intro}
Spherical rolling robots (SRRs) are a fascinating category of robots characterized by their ability to move by rolling on themselves, owing to their unique spherical shape. However, beneath this seemingly simple concept lies a plethora of sophisticated mechanisms and control strategies that enable such motion. Nearly three decades ago, NASA introduced the idea of \textit{``Beach-Ball'' Robotic Rovers}\footnote{\url{https://ntrs.nasa.gov/citations/19950070425}} for planetary exploration, igniting the exploration of various systems in this domain. Notably, the Rollo, designed in 1996 at Finland's Helsinki University of Technology~\cite{halme_motion_1996}, stands as one of the pioneering spherical robots aimed at operating in hostile environments. The inherent protective nature of their spherical shell renders them well-suited for challenging terrains, safeguarding sensitive mechatronics, including sensors and actuators. Consequently, their applications extend to underwater exploration~\cite{lin_development_2012}, surveys of dusty construction sites, tracking crop yields in muddy fields, and even missions in extreme environments such as the moon, Mars, and beyond~\cite{Kalita2020, Rachavelpula2021ModellingRover}. Furthermore, spherical robots have also found utility in educational and therapeutic contexts, particularly for children with developmental disorders~\cite{Mizumura2018MechanicalDisorders}; a market targetted by one of the only companies manufacturing primarily spherical robots, Sphero. 

\begin{figure}[!ht]
    \centering
	\includegraphics[width=\columnwidth]{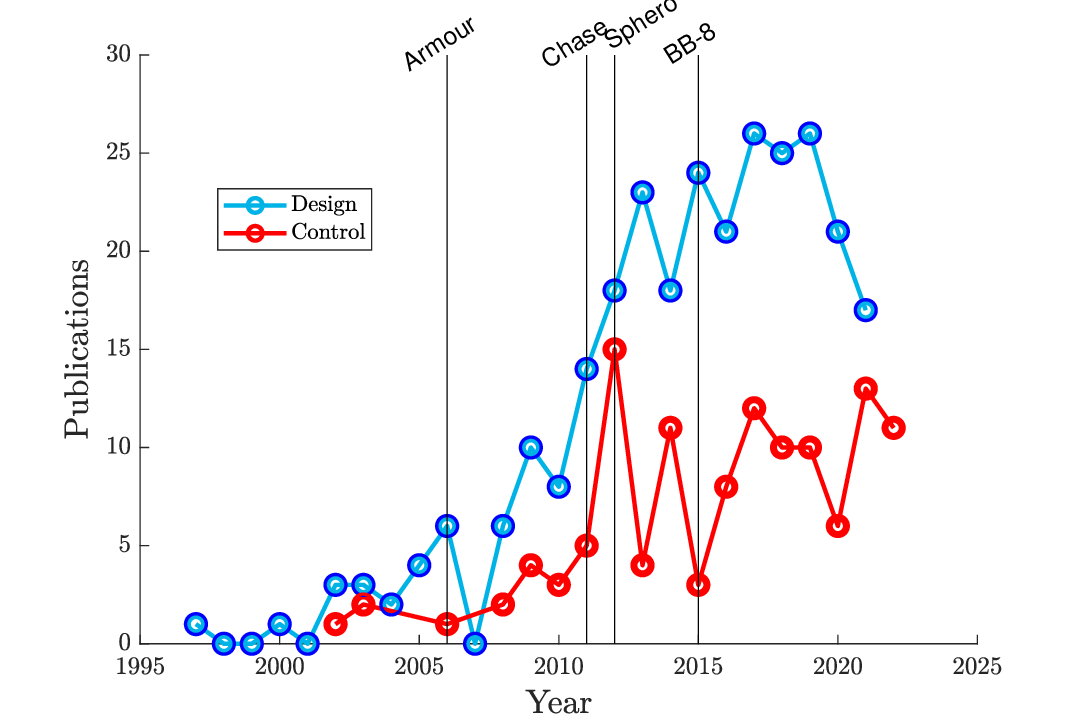}
	\caption{Number of publications each year on the design of SRR (blue) and their control strategies (red). The vertical lines highlight key events in the recent SRR history: the previous literature reviews from Armour (2006) and Chase (2012), the release of Sphero's first product (2011), and the first appearance of BB-8 in Star Wars movies (2015).}
	\label{f:generalSLR}
\end{figure}

Over the years, numerous researchers and companies have proposed diverse designs, dynamic models, and control strategies for spherical rolling robots. While previous reviews exist in the literature, they fail to encapsulate the latest advancements in this field. For instance, a comprehensive examination of rolling in robotics~\cite{Armour2006RollingReview} delves into earlier designs of SRRs, while another review~\cite{chase_review_2012} focuses primarily on the different actuation mechanisms specific to spherical robots. However, since the publication of these reviews, several novel designs have emerged, necessitating an updated analysis (see Fig.~\ref{f:generalSLR}). Although a recent review covered control algorithms~\cite{Karavaev2020SphericalAlgorithms}, it provided limited detail on the employed control strategies due to its concise nature as a conference paper and failed to provide a holistic understanding of both mechanical and control aspects of these robots.

Figure~\ref{f:generalSLR} shows the trend in publication over the period covered in this review (1996-2023) for both mechanical design and control strategies. Since 2010, an increase in interest in SRR is observable. A common misconception is to grant this popularity to the public apparition of BB-8, a fictional spherical robot from the \textit{Star Wars} movie franchise. However, we observe that the trend had started before that, potentially inspiring the filmmakers. We can also observe that \textit{Sphero}'s first product release is right at the start of that new trend, around 2011.

Spherical robots feature a range of internal mechanisms that enable their movement, broadly categorized into three main groups: barycenter offset (BCO), shell transformation, and conservation of angular momentum (CoAM). Barycentric spherical robots (BSRs) manipulate the center of mass to achieve desired motion, exemplified by wheeled mechanisms within a spherical shell or popular pendulum-driven spherical robots. However, the torque capability of BSRs is constrained as the center of gravity cannot be shifted beyond the shell. This limitation can be circumvented through the utilization of control moment gyroscopes (CoAM). Conversely, shell transformation, a less prevalent method, involves deforming the outer body of the robot using wind, air, or water to induce movement~\cite{chase_review_2012}.

The highly nonlinear and non-holonomic nature of spherical robots presents significant challenges in their control. Consequently, various control methods have been proposed in the literature. They most often are designed to be fully autonomous with only a few designed for teleoperation, to the extent of using brain interfaces~\cite{Volosyak_Schmidt_2019, Guan2020RemoteRecognition}. When designed for autonomy, the majority of control strategies can be achieved through nonlinear methods or local linear approximations of the system. This review will show the proportion, advantages and disadvantages of the control strategies found in the literature of SRR, after assessing the underlying dynamic model differences.

In light of the aforementioned landscape, this paper aims to present a systematic review of spherical rolling robots, encompassing the wide range of actuation mechanisms and control strategies implemented in this domain. Through comprehensive data gathering and analysis, the review will shed light on research opportunities and identify blind spots. It is important to note that the scope of this review is limited to robotic systems utilizing their spherical shell for locomotion, as opposed to robots with a spherical shape but employing limbs. Notably, this distinction arises in numerous publications concerning amphibious and underwater spherical robots, where water jets or propellers are utilized in aquatic environments while limbs are employed on land~\cite {Xing2020DesignRobot}. Furthermore, this review focuses solely on active locomotion, excluding spherical robots that rely on external forces, exemplified by the NASA/JPL Tumbleweed polar rover~\cite{behar_NASA/JPL_2004}.

The paper is structured as follows: Section~\ref{s:method} details our methodology, inspired by PRISMA, then Section~\ref{s:locomotion} presents the extracted designs, categorized and compared, followed by Section~\ref{s:control} describing all the control strategies found in our analysis. Finally, in Section~\ref{s:discuss}, we shed light on potential research opportunities and challenges that may unlock the full potential of SRR.

\section{Methodology}\label{s:method}
\begin{table*}[!ht]
    \caption{Keywords prompt designed for each database of this literature review.}
    \centering
    \begin{tabular}{p{0.15\textwidth}|p{0.8\textwidth}}
         \bf{Database} &   \bf{Keywords prompt}  \\
     \hline
        Compendex & (((((design OR control OR command) WN ALL) AND (("spherical Robot*"  OR  ``spherical roll* robot*") WN ALL))) AND ({english} WN LA))\\
    \hline
    Web of Science & (( ( design  OR  control  OR  command )  AND  ( ``spheric* Robot*"  OR  ``spheric* roll* robot*" ) ))
Timespan: All years. Databases:  WOS, CCC, DIIDW, KJD, MEDLINE, RSCI, SCIELO.
Search language=Auto \\
\hline
Science Direct & (Design OR Control OR Command) AND (“spherical Robot” OR  ``spherical Robots" OR “spherical rolling robot” OR ``spherical rolling robots") \\
\hline
Scopus & TITLE-ABS-KEY ( ( ( design  OR  control  OR  command )  AND  ( ``spheric* Robot*"  OR  ``spheric* roll* robot*" ) ) )  AND  ( LIMIT-TO ( LANGUAGE ,  ``English" ) )\\
\hline
ProQuest & ( ( design  OR  control  OR  command )  AND  ( ``spheric* Robot*"  OR  ``spheric* roll* robot*" ) ) \\
\hline
    \end{tabular}
    \label{tab:table1}
\end{table*}

To ensure the utmost rigor and credibility of our analysis and results, we conducted a literature review following the guidelines of the Preferred Reporting Items for Systematic Reviews and Meta-Analyses (PRISMA) framework~\cite{PageM_2021}. Although initially developed for medical and pharmaceutical meta-studies, PRISMA has gained traction and served as a valuable reference for various fields in recent years. In adherence to this framework, our methodology adhered to a comprehensive 27-item checklist encompassing key aspects such as methods, results, and discussions.

\subsection{Publication search and screening}
Our study encompassed all research pertaining to spherical robots, irrespective of publication year, with the caveat that the database content before 1995 is limited. Additionally, we limited our search to English and French publications. To compile our dataset, we performed thorough searches across five information sources: ProQuest, Science Direct, Web of Science, Engineering Village, and Scopus. Employing keywords such as "design," "control," "spherical robots," and "rolling," we identified a total of 3189 records. Subsequently, through a stringent screening process, we excluded records that did not meet the predefined inclusion criteria. As a result, our study included 126 papers concerning control strategies and 280 papers addressing locomotion mechanism design.

During the eligibility assessment, we considered whether the spherical form was an integral aspect of the movement generation, thereby excluding papers where robots relied on external forces for propulsion. This criterion ensured a focused review solely on spherical robots employing their spherical shell for locomotion. To maintain an up-to-date and comprehensive analysis, we encompassed publications from 1996 (no records available before that) up until April 2023.

\begin{figure*}[!ht]
    \centering
	\includegraphics[width=\textwidth]{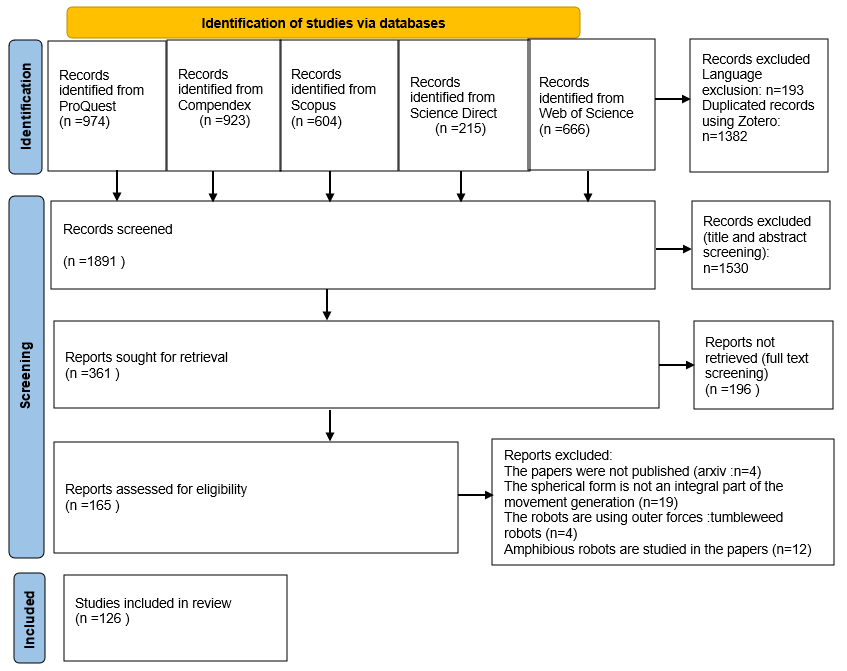}
	\caption{PRISMA 2020 flow diagram for new systematic reviews which included searches of databases~\cite{PageM_2021}}
	\label{f:slr}
\end{figure*}

For each information source, the datasets utilized are presented in Tab.~\ref{tab:table1}, providing transparency and facilitating the replicability of our search process.
The study included 126 papers on control strategies for spherical robots after screening and sorting through 3189 papers obtained from various databases.
The process from the start of the research to the inclusion of the 126 records is summarized in Fig.~\ref{f:slr}. An identical process was applied to the mechanism design topics.

\subsection{Data extraction}
The included records were exported into a software for systematic literature review, which allowed for screening and data extraction. A 7-item form was created for data extraction, including questions about the control strategy used, the objectives of the control method, the outcomes of the method, the driving mechanism of the spherical robot, and whether a simulation or experiment was performed to verify the control strategy:
\begin{enumerate}
    \item Which control strategy is used in the paper? 
    \item Which dynamic modeling technique is used in the paper? 
    \item What are the objectives of the control method? 
    \item What are the outcomes of the method? 
    \item What is the driving mechanism of the spherical robot in the study?  
    \item What sensors have been used or considered? 
    \item Is a simulation or an experiment performed to verify the control strategy? 
\end{enumerate}

Data extraction has been achieved in a redundant manner, in parallel by the first two authors. The presentation and analysis of the content extracted are covered in sections \ref{s:locomotion} and \ref{s:control}.

\subsection{Bibliometric analysis of the keywords}
\begin{figure}[!ht]
    \centering
	\includegraphics[width=\columnwidth]{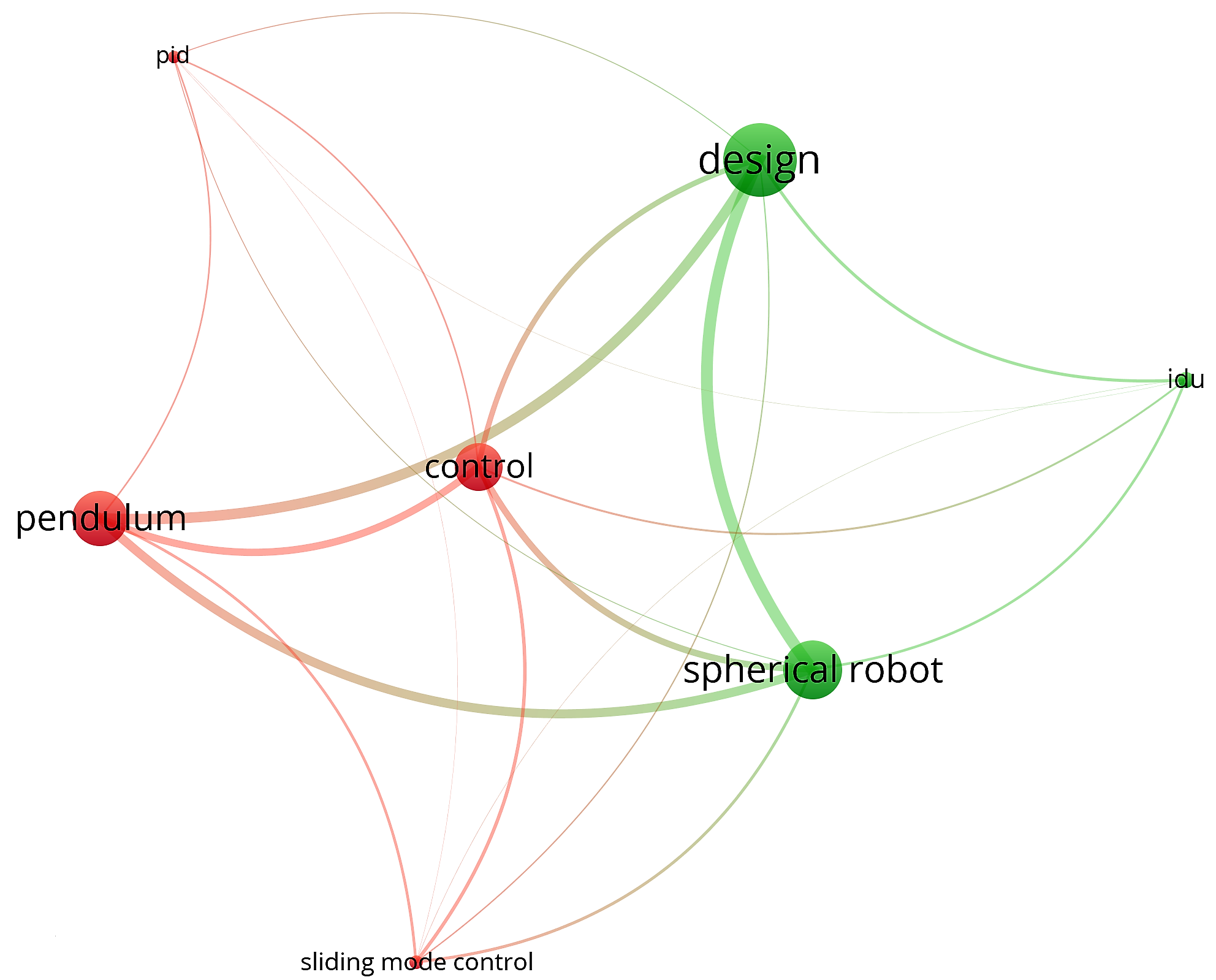}
	\caption{Co-occurrence analysis of the main keywords found in the publications. Made with VOSviewer~\cite{Eck_Waltman_2017}.}
	\label{f:VosViewer}
\end{figure}

To analyze the relationship between keywords, references from various databases were imported into VosViewer~\cite{Eck_Waltman_2017}, which generated a keyword network. The final analysis included keywords that occurred more than 20 times in the papers, and 57 keywords met this threshold. Keywords with a total link strength of less than 600 were excluded from the analysis.  The link strength is the strength of the relationship between the different nodes. The most frequent keywords were "design," "spherical robots," and "control," with total occurrences of 2220, 1841, and 1643, respectively. The size of the node indicated the frequency of a keyword, while the curves represented the co-occurrence of keywords in the same publication. The distance between two nodes determined the number of co-occurrences. The shorter that distance is, the larger the number of co-occurrence will be~\cite{Eck_Waltman_2017}. Figure~\ref{f:VosViewer} shows the different interactions between keywords. The analysis showed that there were more papers on the design of a spherical robot than on its control, and sliding mode control and PID were the most commonly used control strategies. The distance from the spherical robot node to the control and from the spherical robot to the design is the same as from control to design. Therefore, we can conclude that several papers address both control and design aspects, which reinforces the importance of conducting a holistic literature review like this one.

\section{Locomotion, Design and Actuation}\label{s:locomotion}
\begin{figure}[!]
    \centering
	\includegraphics[width=\columnwidth]{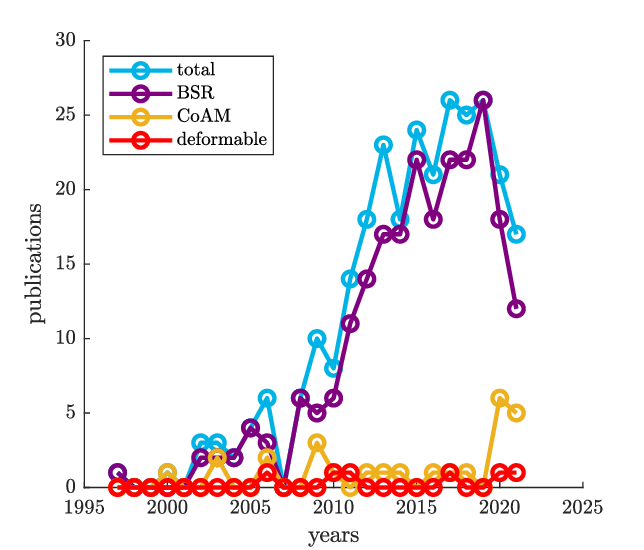}
	\caption{Number of included publications addressing the design of a SRR (some may appear in more than one category, having multiple driving systems)}
	\label{f:DesignYear}
\end{figure}

Our first pass on the large literature gathered focuses on their different actuation mechanisms, which will in turn define the type of system and the control variables. As shown in Fig.~\ref{f:DesignYear}, the number of publications, at least partially addressing the design of the actuation mechanism of an SRR, increased significantly over the last two decades. 

\begin{figure}[!ht]
    \centering
	\includegraphics[width=\columnwidth]{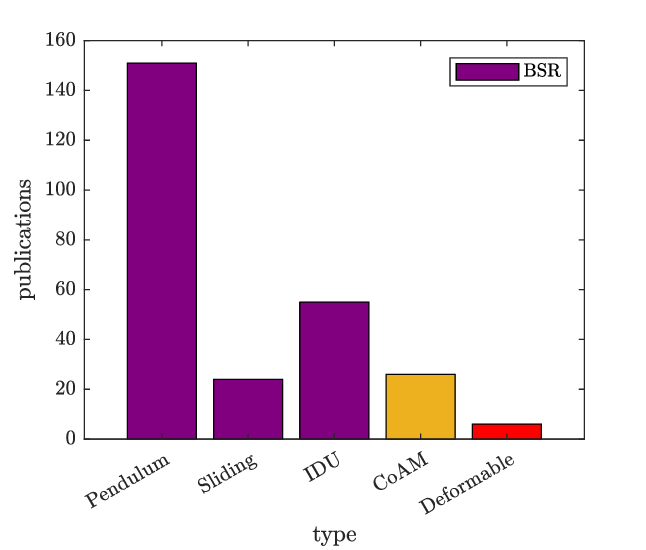}
	\caption{Number of design publications sorted by types of mechanism (robots with more than one actuation mechanism are counted more than once), purple for BSRs, blue for others}
	\label{f:TypeNumber}
\end{figure}

As mentioned above, their locomotion systems can be summarized in three broad categories: \begin{enumerate}
        \item barycentric;
        \item conservation of the angular momentum;
        \item  shell deformation.
\end{enumerate}
Considering these categories, the papers identified and their corresponding devices were classified in Fig.~\ref{f:TypeNumber}. Since an SRR can be actuated with more than one mechanism or with a system fitting in more than one of the above three categories, the sum of robots listed in this figure is higher than the total number of papers addressing the design of the driving mechanism. It should be noted that the BSR category was separated into subcategories in Fig.~\ref{f:TypeNumber}, since it is the most common type, by far. Moreover, colors used in Fig.~\ref{f:DesignYear} correspond to those of Fig.~\ref{f:TypeNumber}.

\subsection{Nonholonomic motion with SRRs}\label{s:holonom}
Holonomic and nonholonomic motion are concepts that are frequently mentioned when it comes to mobile robots. Basically, a nonholonomic mobile robot's state depends on the path taken to reach it. In this case, the velocity constraint defining its motion is non-integral, for instance with ``\textit{rolling without slipping condition}'' with wheels as they cannot slide sideways. Fundamentally, with SRRs, the spherical shape of the shell makes it capable of holonomic motion, as it can roll in any direction. However, it is typically no longer the case when we consider the internal driving mechanism. Indeed, most cannot make their external shell roll in any direction, particularly about their vertical axis. As stated by Chase and Pandya~\cite{chase_review_2012}, for ``true holonomy, the research challenge becomes developing an internal driving mechanism that can provide omnidirectional output torque to a sphere that can arbitrarily rotate around it, regardless of the orientation of either the sphere or the drive mechanism.'' This thus requires an internal driving mechanism capable of moving in three dimensions independently from the spherical shell. According to our review, this is not the case in the overwhelming majority of SSRs, with rare exceptions, which is probably motivated by the fact that holonomic motion is not necessarily needed for most situations, as it is for a car.

\subsection{Barycentric}

By far the most common type of spherical robots according to our assessment, as shown in Fig.~\ref{f:TypeNumber}, barycentric spherical robots (BSRs) are driven by a displacement of their center of mass (CoM). Indeed, by destabilizing the system by moving the CoM away from its point of lowest potential energy (typically directly underneath the center of rotation (CoR) of the sphere), the shell starts rolling. BSRs can be classified into several subcategories:
\begin{enumerate}
    \item pendulum-based;
    \item IDU;
    \item sliding masses.
\end{enumerate}

\subsubsection{Pendulum-based}
Among BSRs, pendulum-based BSRs are the most popular designs. The pendulum typically, but not necessarily, rotates about a shaft passing through the center of the sphere. For a robot to be able to move in more than one direction, at least two DoFs are needed. This can be done with a 2-DoF pendulum, which can rotate in two directions. Otherwise, more than one pendulum can be used. For instance, on the one hand, Li et~al.~\cite{li_spherical_2009} and Zhao et~al.~\cite{zhao_dynamics_2010} proposed BSRs with two, while on the other hand, Dejong et~al.~\cite{dejong_design_2017}, four, to increase their maneuverability. This is obviously done at the expense of more complex control schemes. Moreover, multiple pendulums do not need to be fully independent of each other. For instance, Asiri et al.~\cite{Asiri2019TheRobot} added a second, smaller pendulum orthogonal to the main pendulum rotating about their spherical shell's main axis. Otherwise, to steer its single-pendulum BSR, Schroll~\cite{schroll_angular_2009} designed and patented a differential mechanism to tilt the bob in a direction orthogonal to the rolling motion. While the main axle (passing through the center of the sphere with both extremities rigidly attached to the spherical shell) is usually used as the axis of the pendulum(s), it is not always the case, as demonstrated by Zhao et al.~\cite{zhao_dynamics_2010} which used circular guides carved into the ellipsoid\footnote{For the sake of this review, papers presenting rare rolling robots with slightly ellipsoid shapes were considered, as they behave mostly as spherical robots and their actuation and control scheme can be applied to a perfectly spherical robot} shell of its device. However, having a pendulum only attached to the shell at two places makes it easier to use a flexible shell, as shown by Ylikorpi et al.~\cite {ylikorpi_dynamic_2017}. This characteristic is valuable to overcome small obstacles or steps with a reduced impact on the trajectory, while also attenuating the oscillation due to the nature of a pendulum's motion.
Regardless of the number of pendulums and their designs, the rolling torque of a pendulum-based spherical robot is bounded by the diameter of its shell and the mass of its pendulum. Even if the actuator inside the sphere is capable of generating a larger torque, it will only result in the spinning of the pendulum about its axis, which is not the desired behavior. Moreover, this torque limit is also proportional to the gravitational acceleration. To increase the distance between the center of the sphere and the CoM, many authors located the battery powering their robot in the lowest part of their pendulum, as done by Landa and Pilat~\cite{landa_design_2015}.

While most pendulum-driven robots are incapable of holonomic motion, some have omnidirectional capabilities, notably, the four-pendulum BSR proposed by DeJong et al.~\cite {dejong_design_2017}. In this particular case, this is possible because the four pendulums have four different skew axes intersecting at the center of the sphere.

\begin{figure}[!ht]
    \centering
	\includegraphics[width=\columnwidth]{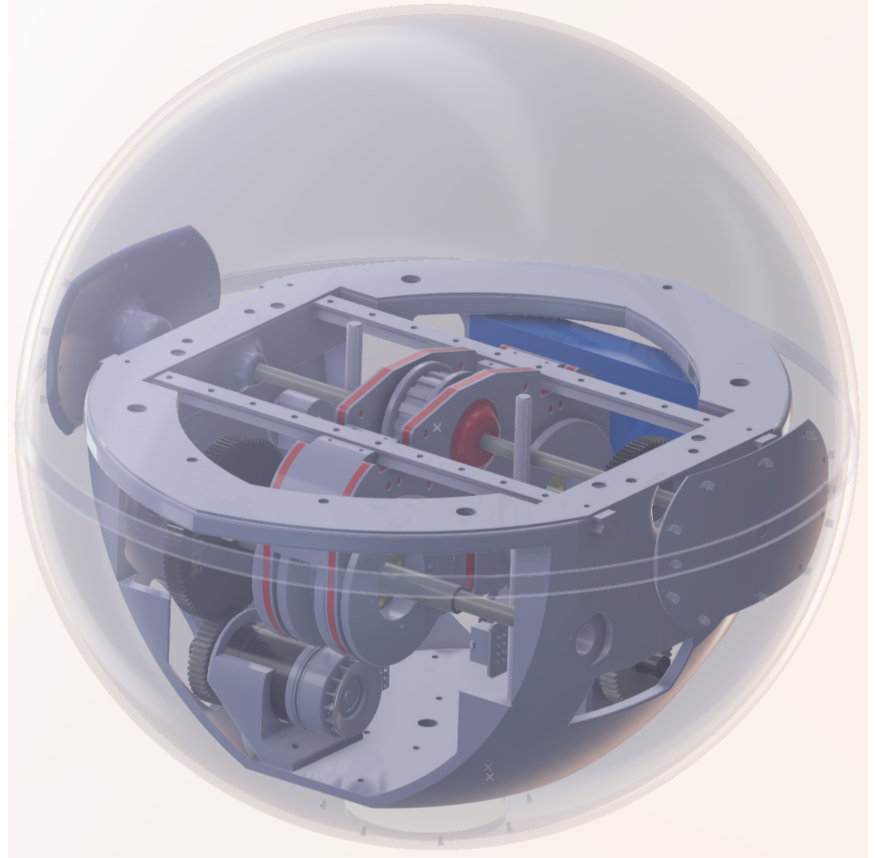}
	\caption{ARIES: a pendulum-based spherical robot     ~\cite{belzile_aries_2022}}
	\label{f:ARIES}
\end{figure}

An example of a pendulum-based spherical robot, known as ARIES, is depicted in Figure~\ref{f:ARIES}. The robot incorporates an actuated cylindrical joint that serves as a pendulum. This novel design allows for simultaneous rolling and steering by utilizing a continuous differential transmission.

\subsubsection{Internal drive unit}
Another type of BSR is those with an internal drive unit. Here, a torque is transmitted to the outer shell through wheels in contact with its internal surface. Sometimes considered separately from IDUs, the \textit{hamster ball} system defines the combination of a spherical shell and a smaller wheeled robot inside the sphere, such as the one proposed by Alves and Dias~\cite{alves_design_2003}. In fact, this type of actuation is among the earliest example of SRR because of its simplicity. Obviously, the potential holonomy of a \textit{hamster-ball} SRR is linked directly to the holonomy of the internal robot. Recently, Belskii et~al.~\cite{Belskii2021DesignLevel} proposed an omnidirectional IDU to be used inside a spherical robot. 
This concept was not fully tested however inside a shell. 

\textit{Hamster balls} suffer from a major issue, namely slipping between the shell and the wheels. To overcome this limitation, IDUs generally incorporate a system to apply a force on the wheel, reducing the risk of slipping. For example, Zhan et~al.~\cite{zhan_design_2011} developed the concept of an omnidirectional IDU with two wheels based on the concept of barycenter offset. Their device consists of a smaller wheel used to orient a larger one in contact with the internal surface of the shell, allowing holonomic motion. Yet, these systems are still known to suffer from slipping between the internal drive and the spherical shell~\cite {chase_review_2012}, limiting their robustness. 

\subsubsection{Sliding masses}
Finally, some BSRs use sliding masses to control the location of the CoM~\cite{mukherjee_simple_1999,bowkett_combined_2017}. For instance, Javadi and Mojabi~\cite{javadi_introducing_2002} used masses moving along prismatic joints to modify the location of the CoM.
However, they are generally, more difficult to control than the two other subcategories described above. Indeed, in the case of masses moving linearly~\cite{javadi_introducing_2002}, a continuous variation of the position of the masses is required, even moving in a single direction, contrary to pendulum-driven robots. This can be overcome by masses moving along circular guides, as Tafrishi et~al.~\cite{Tafrishi2019DesignRobot} did when they investigated the potential of a fluid-based BSR. They used masses moving inside pipes to control the location of the CoM thus generating motion. However, this system was not experimentally validated by the authors.     
As motors can generally be used as generators to accumulate energy, it is also true with spherical robots. Indeed, in applications/environments where minimizing energy consumption is critical, recharging the battery may be useful, as is the case with the robot developed by Zhai et~al.~\cite{Zhai2020ResearchRobot}. The latter harvests energy while the robot is pushed by the wind with electromagnets moving along pipelines. This system can be used to actuate the robot by controlling the position of the magnets instead of letting them move according to the dynamics of the system.

\subsection{Conservation of the angular momentum}
Instead of moving the CoM for the rolling motion, another spherical robot design is to leverage the conservation of the angular momentum without having to displace the CoM. The rolling motion is generated using reaction wheels~\cite{bhattacharya_spherical_2000,joshi_motion_2009,muralidharan_geometric_2015} or control moment gyroscopes (CGM)~\cite{schroll_dynamic_2009,chen_design_2017}. The latter can provide more torque than the former, taking advantage of gyroscopic precession. However, the torque output is not continuous, contrary to the former. Moreover, reaction wheels must be fixed to the shell, rendering the mechanical design and the control challenging~\cite{muralidharan_geometric_2015}. Chase~\cite{chase_analysis_2014} partially overcame this limitation by using four CGMs mounted in dual-scissored pairs.

One of the main differences between BSRs and those based on the conservation of the angular momentum is the non-continuous torque generated for the rolling motion.

\subsection{Shell deformation}
Last but not least, other types of locomotion based on deformable shells are also possible. As can be seen in Fig.~\ref{f:TypeNumber}, they are far less common than barycentric and angular-momentum-based SRRs. This category includes shape-memory alloys and pressurized air bladders~\cite {wait_self_2010}. Their deformable spherical shell can increase maneuverability, as shown by Sugiyama and Hirai~\cite{sugiyama_crawling_2006} with their robot capable of crawling and jumping. Generally able of holonomic motion with a basic controller, they can be complex to design and more prone to failure (i.e. unprotected moving parts).

\subsection{Types and number of actuators}
Most of the mechanisms discussed before can be actuated by different means, namely steppers, servos, brushless, etc. We tried to extract the most common types of actuators from the papers studied, unfortunately, only a few papers provide this level of detail on their prototype. Among them, Chowdhury et al. ~\cite{Chowdhury2018ExperimentsSurfaces} utilized two Faulhaber DC gear drive motors, each with 30mNm/3V, to control a 90-gram spherical robot. Similarly, \cite{belzile_aries_2021} used two Maxon EC 45 flat 30W brushless motors with corresponding controllers to actuate a 10kg spherical robot. They justified this choice based on the motors' lightweight nature and their ability to fit within the limited available space.
In contrast, Azizi and Keighobadi \cite{Azizi_Keighobadi_2014} proposed a design with three motors driving three independent rotors. Similarly, Jia et al. \cite{jia_motion_2009} proposed a design with three independent actuators, including one drive motor, one steer motor, and one flywheel. The spinning flywheel enabled increased angular momentum of the spherical robot. In  \cite{Roozegar2017ModellingResults}, two stepper motors were used as actuators, where one rotated the main shaft and the other enabled the rotation of the pendulum. Additionally, four gears were utilized for power transmission. In a different approach,  \cite{Zheng2021ResearchMethod} employed three separate motors in the internal mechanism: a flywheel motor for increasing angular momentum, a long-axis motor for generating driving forces, and a short-axis motor for controlling the counterweight's angle relative to the shell's axis.

\subsection{Design scale}
\begin{figure}[!ht]
    \centering
	\includegraphics[width=\columnwidth]{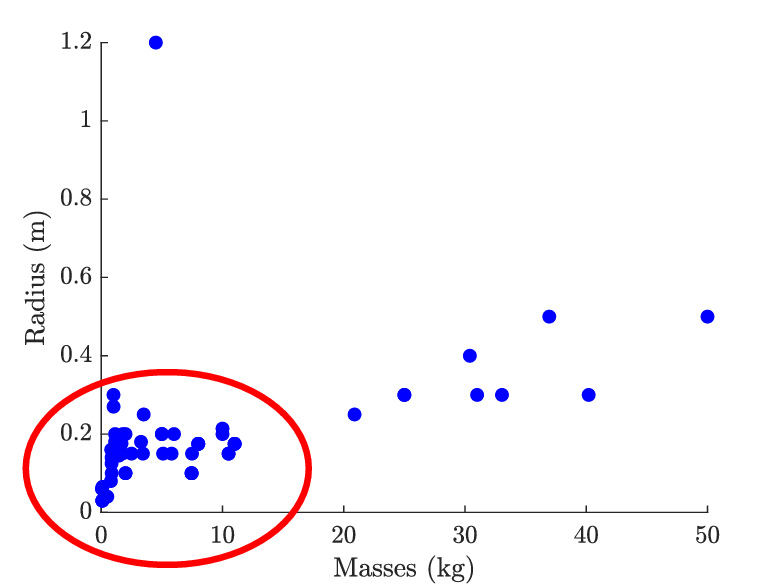}
	\caption{Radius and mass of included mechanisms}
	\label{f:Rayon}
\end{figure}

The papers examined in this study featured spherical robots of different scales, e.g. weight and dimensions. Figure~\ref{f:Rayon} visually represents the masses and radii of the described robots, where such information was available. It is worth noting that some authors, as observed in~\cite{Yue2012DisturbanceTechnology, Yue2014ExtendedRobot}, employed the same spherical robot in multiple papers. To ensure accuracy, we considered each robot only once, resulting in a total of 58 distinct robots represented in the figure.

Among the examined robots, the lightest variant, proposed in~\cite{Niu2014MechanicalRobot}, weighed a mere 55 grams, while the heaviest robot, documented in~\cite{Zhai_Li_Luo_Zhou_Liu_2015}, weighed a substantial 50 kilograms. The analysis revealed that the majority of the examined robots weighed less than 10 kg, which can be attributed to their primary usage in exploration scenarios. The benefits of lighter robots are twofold: they require less energy to operate and are more manageable for transportation purposes.

Furthermore, the investigation into the papers unveiled that the majority of the robots possessed a radius ranging between 0.1 and 0.2 meters, with the largest radius belonging to the device proposed by Rigatos et al.~\cite{Rigatos2019NonlinearRobot}, measuring 1.2 meters. In contrast, the smallest robot introduced by Nguyen et al.~\cite{Nguyen2017}, exhibited a radius of 30~cm. At this scale, the onboard computer and sensing capacity are inherently limited, imposing constraints on the robot's capabilities.

\section{Control and dynamics}\label{s:control}
\begin{figure}[!ht]
    \centering
	\includegraphics[width=\columnwidth]{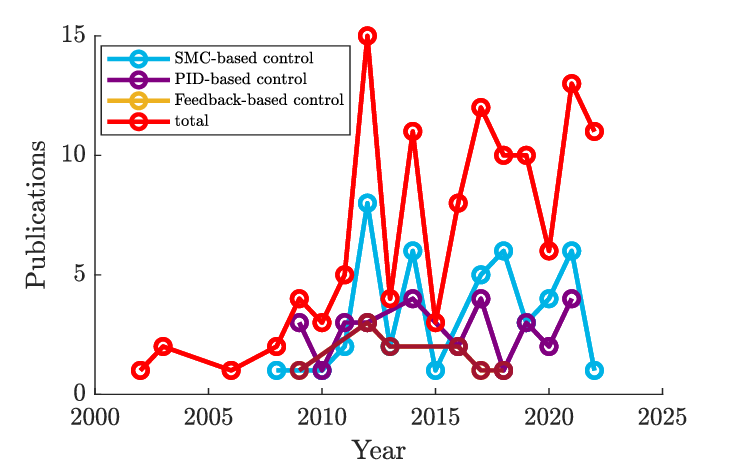}
	\caption{Number of papers addressing the control of a SRR following the most used control strategies.}
	\label{f:ctrpapers}
\end{figure}

During our thorough analysis of the wide array of publications encompassed in this study, our subsequent phase focused on an examination the diverse range of control strategies. In this context, Fig.~\ref{f:ctrpapers} illuminates a significant trend: the emergence of publications focusing on the control of spherical robots began gaining prominence around 2012, culminating in a notable peak in 2019.

\subsection{Dynamic modeling}

Spherical robots possess a distinctive motion mechanism, resulting in intricate kinematics and dynamics to model and analyze. The incorporation of non-holonomic constraints, as discussed in Section~\ref{s:holonom}, further amplifies the complexity of these models. While certain papers, such as~\cite{Nam2019DeepRobot, Li2018, Li2019Kinematics-basedPendulum}, rely solely on the kinematic model for controlling spherical robots, most studies employ a dynamic model due to the influence of conservative forces, such as the pendulum gravity effect. However, when feasible, the kinematics-only control approach offers a simplified depiction of the system's behavior, emphasizing motion or positional characteristics rather than intricate dynamics.

The dynamics of a spherical robot can deviate from those of traditional robots due to factors such as the shape of the spherical body, ground interactions, and force or torque distributions, all of which impact its motion. Developing a dynamic model is a crucial step prior to delving into the control of robotic systems. In this research, various methods were employed to model spherical robots.

\begin{figure}[!ht]
    \centering
	\includegraphics[width=\columnwidth]{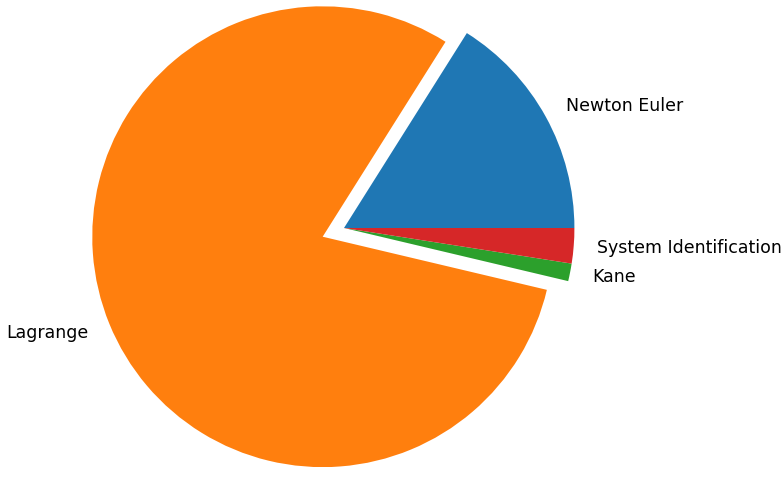}
	\caption{Dynamic modeling methods found in SRR papers from 1996 to 2023. The Euler-Lagrange method largely predominates the others.}
	\label{f:dynamic}
\end{figure}

Figure~\ref{f:dynamic} illustrates the most commonly utilized techniques for modeling spherical robots. The Euler-Lagrange method is extensively employed in numerous papers for deriving the dynamic model. This method provides a systematic approach by considering the system's energy and serves as a valuable tool in modeling mechanical systems. However, the presence of motion constraints may require the use of Lagrange's multipliers, which in turn adds an additional computational burden for calculating the corresponding constraint forces.
The Newton-Euler method, widely utilized in robotics, physics, control systems, and other fields requiring accurate dynamic modeling, offers explicit equations of motion and a recursive formulation, enabling detailed analysis and control of complex mechanical systems. Its explicit representation of inter-body forces facilitates structural analysis of all the components. 
Alternatively, system identification techniques can be employed to create mathematical models describing a system's behavior based on input-output relationships, rather than relying on a physics-based representation. This successful strategy has been implemented in~\cite{Kamaldar2011ARobot, Kamaldar2011RobustModeling}, eliminating the need for explicit knowledge of the underlying physical equations governing the system. However, system identification has its limitations, such as the requirement for high-quality data, the potential for modeling errors if the data is unrepresentative, and the need for careful experimental design.

In~\cite{Azizi_Keighobadi_2014}, the Kane method was utilized to obtain the dynamic model of a spherical robot. Although this method may be more complex to implement compared to simpler approaches like the Newton-Euler method, it offers significant advantages in terms of precision and representation, particularly for complex mechanical systems or detailed analyses. 
In \cite{Sandino_Bejar_Ollero_2013} a comparison of the different methods for modeling a Small-size Helicopter has been done. The results showed that Kane's method offers unique advantages compared to traditional approaches. It incorporates constraints using generalized coordinates and speeds, resulting in concise equations. Constraint forces are considered from the start, and the method is known for its computational efficiency.
However, implementing Kane's method necessitates a solid understanding of multibody system dynamics.

\subsection{Control strategies}
A total of 33 control strategies were identified in this study, with sliding mode-based control, PID, PD, fuzzy control, and feedback linearization being the most prevalent. However, these strategies are often enhanced through the incorporation of additional methods, such as fuzzy control or neural network algorithms, into traditional control approaches. Other strategies, such as PD, feedback linearization, and backstepping, are also reported.

\begin{figure}[!ht]
    \centering
    \includegraphics [width=\columnwidth]{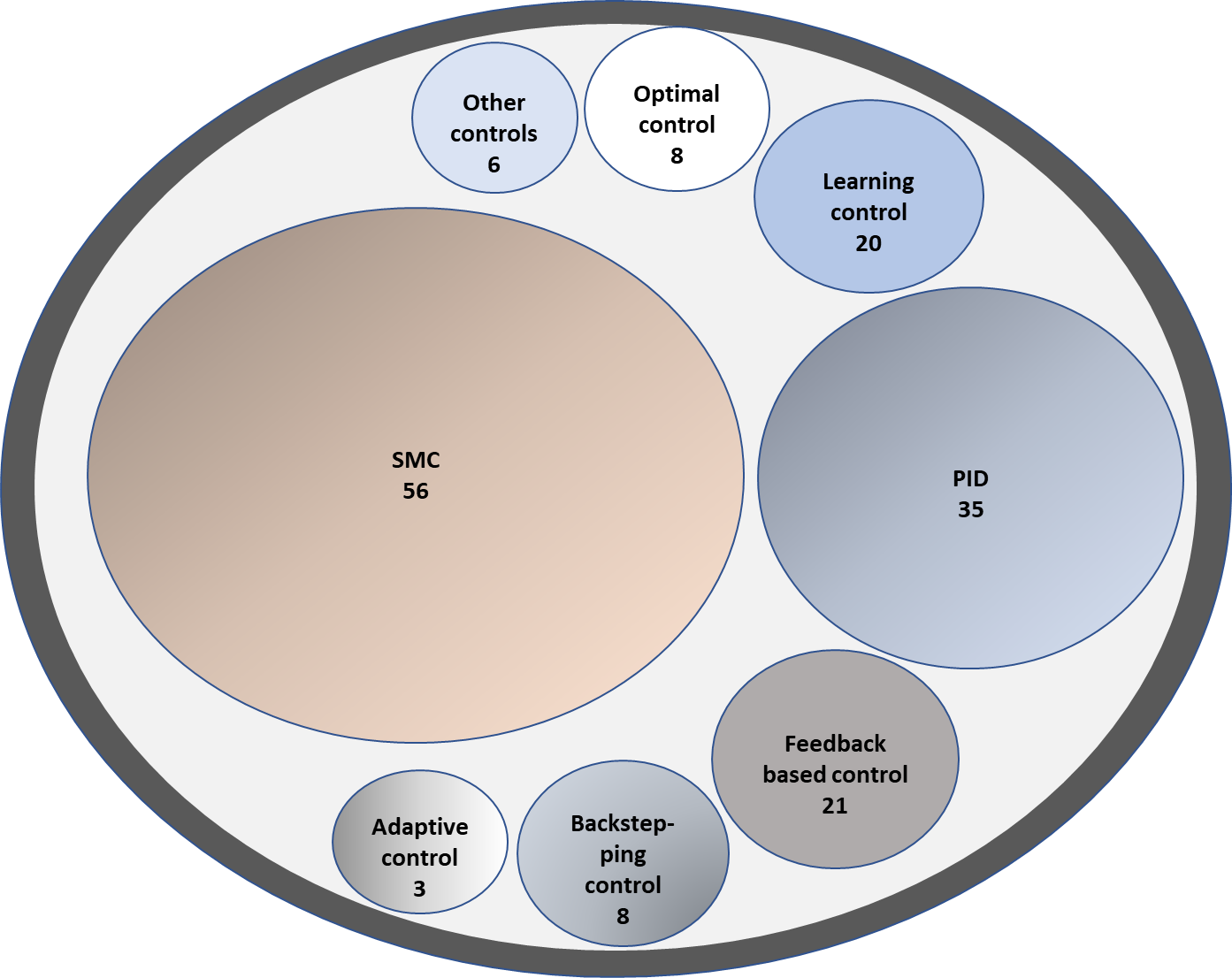}
     \caption{Most prominent identified control strategies and their relative importance from the number of related publications. 'others control' are detailed in Fig.~\ref{f:control}.}
    \label{fig:c_strat}
\end{figure}

This section provides a comprehensive listing and detailed explanation of the various control strategies. Some strategies combine two or three control methods, while other papers independently investigate multiple control strategies for the purpose of comparison. Figure~\ref{fig:c_strat} provides a summary of the different control strategies documented in the papers.

Among the designs used, the barycentric spherical robot design was the most commonly employed, with sliding mode-based control being the preferred control strategy. However, no direct correlation between the design of the spherical robot and the chosen control strategy was observed.

\begin{figure*}[!ht]
    \centering
	\includegraphics[width=\textwidth]{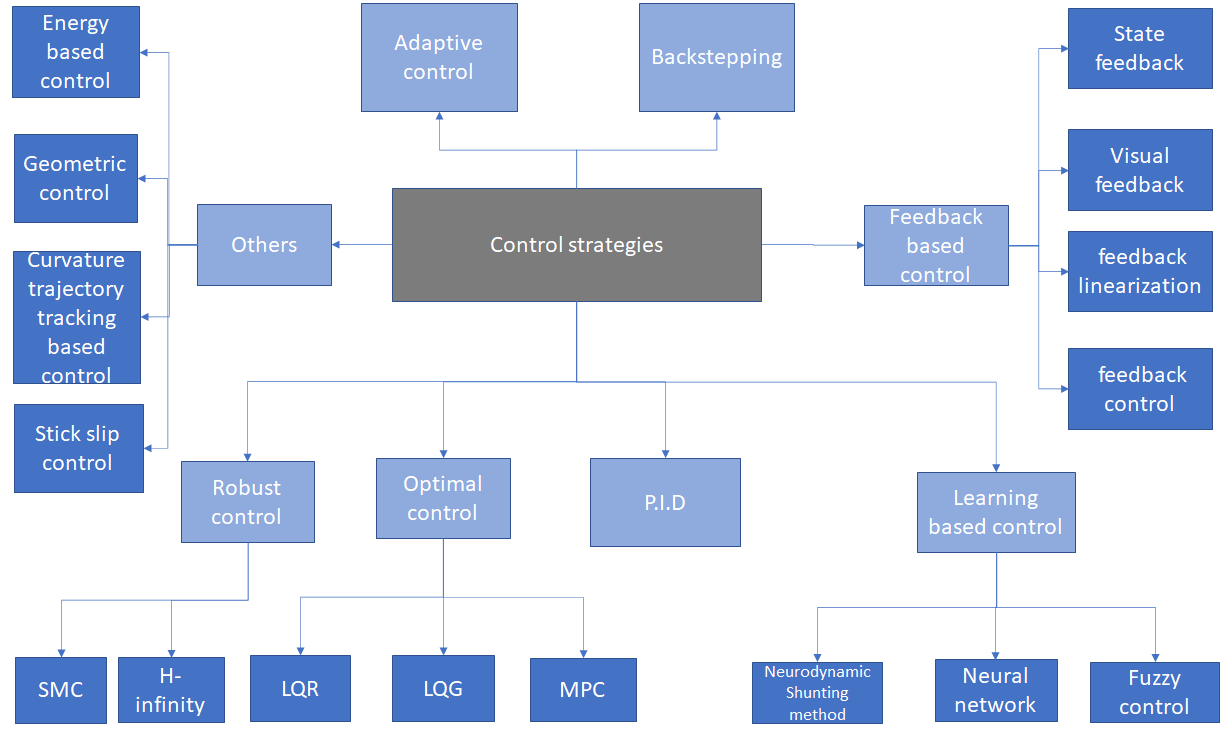}
	\caption{Different control strategies included in the study}
	\label{f:control}
\end{figure*}

For an overview of the diverse control strategies included in this study, please refer to Fig.~\ref{f:control}.

\subsubsection{Sliding mode control}\label{SMC}
Sliding mode control is a widely employed technique for controlling various types of robotic systems, including spherical robots and manipulators. Its effectiveness in handling uncertainties and unmodeled disturbances, along with its robust performance, has made it a popular choice. Numerous studies have proposed different SMC-based control strategies, such as cascade sliding mode control (CSMC) \cite{Huang2012CascadeWheels}, high-order sliding mode control (HSMC) \cite{Chowdhury2018ImplementationRobot}, and adaptive hierarchical sliding mode control (AHSMC) \cite{Yue2014ExtendedRobot}, to enhance the robustness and stability of spherical robots. Researchers have also explored the combination of SMC with adaptive control laws, neural networks, and fuzzy control to address uncertainties and reduce chattering. These hybrid approaches have demonstrated improved tracking accuracy, convergence speed, and robustness compared to traditional SMC. Fuzzy control, in particular, is well-suited for complex robotic systems with challenging modeling requirements.

The selection of SMC in the included papers can be attributed to several reasons, primarily the desire to achieve enhanced robustness and cope with uncertainties and unmodeled disturbances. This rationale is discussed in~\cite{Liu2008StabilizationRobot, Liu2010NonlinearRobot, Wang2011ConstantWheels, Zheng2011ControlDynamics,Yu2013PathPlane, Azizi_Keighobadi_2014,Zhao2014ARobot, Ayati2017,Li2020Anti-disturbanceRobot, Nguyen2017}.

One promising implementation strategy involves decomposing a complex robotic system into simpler subsystems, where a combination of the dynamic model states and sliding surfaces, known as cascade sliding mode control (CSMC), can be applied as demonstrated in \cite{Huang2012CascadeWheels}. However, conventional CSMC faced challenges in achieving the desired position control performance due to switching constants. To overcome this, the authors introduced positive constants and referred to this modified control method as CSMC1.

Typically, the dynamic model of robots is necessary for designing a sliding mode controller. However, in \cite{Li2018}, the authors introduced a kinematic-based SMC that can simultaneously track the position and attitude states of a spherical robot. To reduce chattering, a saturation function was added to the traditional SMC approach.

In the pursuit of stability in robot control, \cite{Chowdhury2018ImplementationRobot} proposed both first and second-order SMC methods. They demonstrated that high-order sliding mode control (HSMC) outperformed the first-order approach, a finding corroborated in \cite{Bastola2019SuperRobot}. The control strategy presented in the latter paper reduced chattering and enhanced robust operation in the presence of disturbances.

For improved performance in the face of uncertainties, in~\cite{Ma2020Fractional-OrderRobot} the authors introduced a fractional-order adaptive integral hierarchical sliding mode control. This approach combines adaptive control laws, hierarchical sliding mode control, and an integrator in the control loop. Comparative analysis with adaptive hierarchical sliding mode control and HSMC revealed that the proposed method exhibited superior convergence speed, stability, and robustness. Adaptive hierarchical sliding mode control was initially proposed in \cite{Yue2012DisturbanceTechnology} and \cite{Zhang2021BalanceAsymmetry}, while HSMC was presented by authors such as \cite{Cai2012PathRobot} and \cite{Chiu2012HierarchicalWheels}.

In \cite{SalemizadehParizi2021}, the authors introduced a hybrid super-twisting fractional-order terminal sliding mode control for a rolling spherical robot, which demonstrated higher accuracy and reduced chattering compared to traditional SMC and other SMC-based controls. Hybrid super-twisting fractional-order terminal sliding mode control is an advanced control technique that combines super-twisting control, fractional-order control, and terminal sliding mode control. It aims to achieve robust and accurate control of dynamic systems, especially nonlinear and uncertain systems.

To stabilize a spherical robot moving on an inclined plane, in ~\cite{Roozegar2017MathematicalControl} the authors presented a terminal sliding mode control. They showed that this control approach effectively reduced chattering compared to SMC and facilitated rapid convergence of the tracking error to zero in less than one second, outperforming SMC for their specific spherical robot.

Some authors have integrated adaptive laws with SMC to enhance results. Adaptive control enables the modeling of uncertainties, thereby reducing chattering~\cite{Chowdhury2018}. In \cite{Sadeghian2015}, an adaptive method was employed to identify the exact model. Similarly, in \cite{Yue2014AdaptiveResistance} the authors employed the same approach to improve robustness and suppress vibrations in the inner suspension platform of a spherical robot. Hierarchical sliding mode control with adaptive methods, referred to as adaptive hierarchical sliding mode control (AHSMC), not only offers improved robustness and parameter insensitivity characteristics but also estimates dynamic disturbances \cite{Yue2014ExtendedRobot}. In \cite{Zhang2021BalanceAsymmetry}, AHSMC was utilized for estimating the moment and rolling friction of a spherical robot.

Authors of \cite{Chen2020RecurrentRobot} combined SMC with neural networks to achieve better performance. The former proposed a radius-based function for controlling the forward movement of the robot, demonstrating good convergence. The latter employed recurrent neural networks in conjunction with SMC (RNNSMC) for stabilization and tracking control. This approach improved the accuracy and performance of the model, with comparisons against fuzzy PID indicating its effectiveness in reducing chattering. Fuzzy PID was chosen as a baseline due to the performance improvements offered by neural network-based control and the suitability of fuzzy control for complex robot modeling.

We note from the above studies that SMC alone often falls short in adequately addressing uncertainties and modeling errors in SRR systems. To tackle this limitation, researchers have also explored the integration of neural networks and fuzzy control with SMC to achieve improved robustness and faster convergence. For instance, in \cite{Kayacan2013AdaptiveAlgorithm}, the authors incorporated neural networks and fuzzy control into the SMC framework, yielding higher robustness and faster convergence in the presence of uncertainties. Similarly, in \cite{Andani2018Fuzzy-BasedRobot}, fuzzy control was combined with SMC to effectively handle parameter uncertainties. This integration of fuzzy control with SMC proved instrumental in enhancing the overall performance of SRR systems.
\subsubsection{PID, PI, and PD control strategies}
PID control is widely employed by researchers in their studies, either in its traditional form or in combination with other control laws such as fuzzy control, neural networks, Cerebellar-model articulation control, or geometric and adaptive laws. It is probably the most intuitive and less demanding, in terms of processing power, controller for its performance. It is thus common to select this strategy for a fast deployment. For instance, the traditional PID control is utilized in \cite{Jayoung2009AImplementation} to demonstrate its functionalities. In the work by \cite{Roozegar2017ModellingResults}, PID control is proposed, following Ziegler-Nichols rules, to stabilize a spherical robot on an inclined plane. The Ziegler-Nichols rules are a set of heuristic guidelines used in control theory to tune the parameters of a proportional-integral-derivative (PID) controller.
The Authors in \cite{Li2020Anti-disturbanceRobot} apply PID control for pitch angle control. Furthermore, in \cite{Cai2011NeuralRobot}, PID control is combined with neural networks to compensate for actuators' nonlinearity when single input multiple output PID control fails. This integrated approach exhibits fast convergence and proves to be an efficient method for controlling SRRs.

The application of fuzzy PID control is proposed in \cite{Roozegar2017ModellingResults} for the motion of a spherical robot, while a comparison with PID control in \cite{Sadeghian2016ControllerRobot} demonstrates that fuzzy PID control outperforms PID in terms of performance and stabilization time. In \cite{kayacan_velocity_2012}, fuzzy PID control is combined with feedback linearization and a grey predictor, with the output of the grey predictor utilized instead of the current system outputs. This method exhibits superior performance compared to PID and fuzzy PID controls in terms of overshoot and settling time. Moreover, some studies employ the proportional-integral (PI) control strategy. For example, \cite{Belskii2021} proposes PI control for controlling the angular velocity of a spherical robot, and \cite{Tang2021Co-SimulationMATLAB} employs PI control for velocity control of a two-wheel differential spherical robot. In \cite{Zhang2009ApplicationRobot}, PI control is employed for real-time control of a spherical robot, with the addition of a genetic algorithm to enhance stability.

PD control is also utilized in certain papers, such as \cite{Hanxu2010DymanicsRobot} and \cite{jia_motion_2009}, where it is proposed for velocity control of an omnidirectional robot with flywheels. In \cite{Nguyen2017} and \cite{Niu2014MechanicalRobot}, PD control is employed for controlling the yaw orientation of the spherical robot, while no other strategies are presented for controlling the remaining degrees of freedom.

\subsubsection{Other control strategies}
In addition to sliding mode control (SMC) and proportional-integral-derivative (PID) control, various other control strategies have been employed to control robotic systems. For instance, backstepping control combined with neural networks was used in \cite{Ghommam_Mahmoud_Saad_2013} to achieve high-performance cooperative control of a group of mobile robots. The linear quadratic regulator (LQR), dynamic programming, and model predictive control have also been utilized to control spherical robots in some of the included papers. For example, LQR has been used to control robots or stabilize motion, as in  \cite{zadeh_lqr_2014}, while dynamic programming has enabled optimal motion planning and control of spherical robots in \cite{Roozegar_Mahjoob_Jahromi_2016}. Feedback linearization has been proposed in  \cite{Kayacan2012ModelingApproach, Huang2017DynamicalWheels}  to achieve control and stabilization objectives, and in  \cite{Hanxu2010DymanicsRobot}, the linear quadratic regulator is used in combination with percentage derivative control for better performance. Some control methods have been used only once, such as the active disturbance rejection control in \cite{Zhou2021} to achieve trajectory tracking control for biomimetic spherical robots without model parameter information, energy-based control in \cite{Sun2021ARobot} for a spherical robot to track high moving speeds, and H-infinity control in \cite{Rigatos2019NonlinearRobot} 
 to provide a solution to the robot's optimal control problem under model uncertainty and external perturbation.

\section{Experiments and sensing}

\subsection{Simulations vs experimental validation}
In all included papers, authors performed either experiments or simulations in order to validate or verify the feasibility of the proposed control strategies. From the control aspect, 32~\% of the papers performed experiments to validate their control strategies while 78~\% realized only simulations. These validations were done in numerical simulation software such as Matlab Simulink~\cite{Cai2012PathRobot} or by implementing the control strategy on a prototype~\cite{jia_motion_2009}. If we look only at BSRs, the broad mechanism category with the largest number of publications over the years, a significant portion of them included a physical prototype, namely 50~\% for pendulum-based BSRs, 38~\% with sliding masses and 82~\% for IDU-driven BSRs. These numbers are not unexpected, as robots with an IDU are generally cheaper to prototype. On the opposite end, using sliding masses increases the mechanical complexity of the system, as well as requiring more complex control schemes, the sliding masses needing to continuously change direction to provide a uniform and continuous motion of the sphere. 

Beyond the presentation of a prototype, experimental validation with different tests and scenarios is less common.
For instance, Chowdhury et~al.~\cite{Chowdhury2018ExperimentsSurfaces} conducted experiments in both indoor (smooth surface) and outdoor environments (rough surfaces) to validate their model and their control strategy as well. In the paper by Alves and Dias~\cite{alves_design_2003}, a path curvature-based control was presented.
Some authors used motion capture systems able to track markers inside the shell in order to validate the accuracy of robot localization \cite{Ajay_Suherlan_Soh_Foong_Wood_Otto_2015}.
When conducting experiments, different metrics were selected for the validation of the results. The most common were the overshoot, the steady state response, the rise time, the convergence time, and the maximum absolute value of motion trajectory error.

\subsection{Embedded sensing}
The few reviewed papers that progressed to prototype validation necessitated the incorporation of sensing components to close the control loop. The majority of these papers employed commonly used sensors, including accelerometers, cameras, and encoders. Accelerometers, typically housed within an inertial measurement unit (IMU), were utilized in some papers to acquire feedback on the robot's state. Notably, \cite{alves_design_2003} proposed to use a three-axis accelerometer combined with a three-axis rate gyroscope to obtain information about the robot's acceleration, angular velocity, and position. In \cite{zhao_dynamics_2010}, the feedback is obtained from a module of sensors that consists of encoders of the servomotors, angular rate sensors, angular acceleration sensors, and directional gyro, which are all microelectromechanical systems (MEMS) devices However, IMUs were predominantly deployed to provide orientation information in studies such as \cite{alves_design_2003, Niu2014MechanicalRobot, Zhai_Li_Luo_Zhou_Liu_2015, Zhang_Ren_Zheng_2022, Zhang_Ren_Zheng_2023, Zheng2021ResearchMethod}. To mitigate noise and drift issues, an extended Kalman filter was applied to the IMUs in \cite{9674842}.

Some researchers incorporated additional sensors into their systems. Encoders were commonly used in conjunction with IMUs \cite{Liu_Sun_Jia_2009, Urakubo2016DynamicGyro, Urakubo2012DevelopmentGyro, Ajay_Suherlan_Soh_Foong_Wood_Otto_2015, Chowdhury2017, Hu_Guan_Lin_Wang_Liu_Wang_Li_2022, Ling_Zhang_Weng_Cai_Li_Zhang_Xiao_2022, RoyChowdhury2017ExperimentsProject, Wang_Guan_Hu_Zhang_Wang_Wang_Liu_Li_2021, Wang_Liu_Guan_Hu_Zhang_Wang_Hao_Li_2023}, while cameras were employed for real-time position tracking in \cite{Zhang_Ren_2022, Zhang2021BalanceAsymmetry}. In \cite{Hanxu2010DymanicsRobot}, an IMU, gyroscope, and inclinometer were combined to acquire the pitch, roll, and yaw angles, as well as the lean angles and spin rate of the robot. To improve the accuracy of position measurement considering slipping, \cite{Cai2012PathRobot} proposed a vision system in addition to an IMU. Vision systems were utilized by other authors for obtaining feedback information as well. For example, \cite{Jaimez2012DesignRobot} proposed a system comprising four infrared cameras along with dedicated software to determine the position of the robot in space. In \cite{Roozegar2018AdaptiveSystem}, an RGB camera is used to track the spherical robot pose at 30Hz (each frame). Similarly, in \cite{Roozegar_Mahjoob_Jahromi_2016, Sakalli_Beke_Kumbasar_2018}, a camera system was utilized to acquire the robot's position. Furthermore, in \cite{belzile_aries_2021}, the combination of a depth camera and a LIDAR was used to map the environment and infer the robot's position in the map (Simultaneous Localization and Mapping - SLAM). It should be noted that employing camera systems either requires a transparent shell or a turret to house the camera, resembling the popular design of BB-8.

\begin{figure}[!ht]
    \centering
	\includegraphics[scale=0.27]{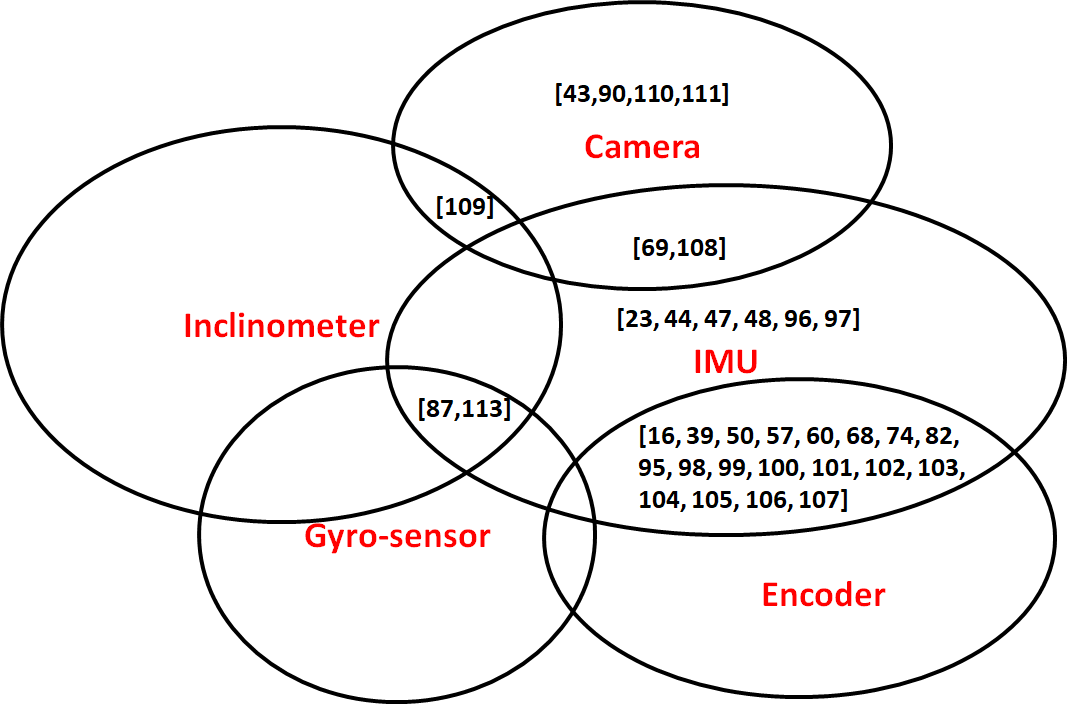}
	\caption{Embedded sensors distribution and combination in the publications covered. The vast majority combine IMUs with encoders.}
	\label{f:s}
\end{figure}

An overview of the sensors used in the reviewed papers is depicted in Fig.~\ref{f:s}.

\section{Discussion}\label{s:discuss}
After a comprehensive analysis of the papers, several challenges associated with spherical robots have been identified. While some of these challenges have been addressed in the literature, others remain unexplored. In this section, we will delve into these challenges and discuss potential avenues for further research.

One significant challenge arises from the limited number of contact points that spherical robots have with the ground. This constraint makes it challenging to employ traditional sensors similar to those used in wheeled or legged robots for measuring motion and orientation. Although some papers suggest using IMUs for measurements, controlling the robot becomes problematic due to drift errors inherent in IMUs. Consequently, there is a research opportunity to explore robust control methods that can effectively model disturbances and uncertainties. However, relying solely on proprioceptive measurements may prove insufficient for most realistic missions such as exploration and inspection.

To overcome the limitations of proprioceptive measurements, additional sensors like cameras, LIDAR, or ultrasonic sensors can be integrated to complement IMU data and provide more accurate information about the robot's surroundings and motion. An example of such integration can be found in \cite{belzile_aries_2021}, where a camera and a LIDAR are utilized. However, the use of cameras and LIDARs in spherical robots introduces challenges related to refraction and distortion. These phenomena can alter distance and position measurements captured by the sensors, leading to perceptual errors. Moreover, inadequate perception of the environment's geometry can result in errors in motion planning and localization. Thus, investigating the impact of refraction and distortion becomes crucial in enhancing perception precision, motion planning, localization, and sensor calibration. Such research efforts would ultimately improve the reliability and efficiency of navigation for robots equipped with cameras and LIDARs.

When considering the integration of cameras and LIDARs, it is imperative to ensure a clear line of sight. Alternatively, cameras can be positioned outside the shell on both sides, as demonstrated in \cite{Kolbari2018ImpedanceRobot}. However, integrating cameras outside a non-transparent shell may impose limitations on the robot's omnidirectional movements and increase the risk of camera damage.

The unique shape and locomotion mechanism of spherical robots poses significant challenges in obstacle avoidance. These robots rely on changing their center of gravity or rolling in different directions to move. However, the lack of independent directional control makes efficient navigation around obstacles challenging. Additionally, SRRs maintain contact with the ground or surfaces using a single point or a limited set of points, which can affect their stability and maneuverability, especially when encountering uneven terrain or obstacles with varying heights. Maintaining balance and stability while avoiding obstacles becomes increasingly challenging in such scenarios. One potential solution to address this challenge is incorporating jumping mechanisms in spherical robots, enabling them to overcome obstacles vertically. By introducing a jumping mechanism, spherical robots gain the ability to leap over barriers, gaps, or rough terrain that would otherwise be challenging or impossible to traverse with rolling or sliding locomotion alone. However, achieving controlled leaps while managing the position of the center of mass and the point of application of the propulsion force remains an open challenge due to the mechanical complexity involved.

Moreover, two distinct approaches can be employed to tackle the intricate dynamics of spherical robots: the decoupled approach and the complete approach. The decoupled dynamic approach involves separating the transversal and longitudinal motions of the robot and studying them independently. This approach simplifies the dynamics and reduces computational effort. On the other hand, the complete dynamic approach involves modeling the robot without decoupling the motions, which adds complexity to the system. Comparing these two approaches can be valuable in understanding their respective advantages and drawbacks, enabling individuals to determine the most suitable approach for their needs.

In addition to the aforementioned challenges, controlling a spherical robot can be particularly demanding, especially for non-holonomic systems such as pendulum-driven spherical robots. These non-holonomic systems possess motion constraints that cannot be easily integrated into their configuration space. As a result, designing control laws capable of steering the system along a desired trajectory becomes difficult. Unlike holonomic systems, non-holonomic systems require careful consideration of their constraints and dynamics to develop control laws that can achieve the desired motion. Therefore, further research is needed to explore effective control strategies tailored to non-holonomic spherical robots.


\section{Conclusion}\label{sec:conc}

The paper provides a comprehensive analysis of spherical robots, covering various aspects such as design, locomotion, control, and embedded sensing. It reviews multiple research papers to gain insights into the state of the art in spherical robot development.
The design of spherical robots involves considerations such as the mechanisms for locomotion. Different locomotion mechanisms are discussed, including pendulum-driven systems, internal mass shifting, COAMs, and shell transformation.
The paper explores various control strategies for spherical robots, with a focus on efficiency, stability, robustness, and feasibility. A comprehensive analysis identified a total of 33 control strategies used for controlling spherical robots. Among these strategies, the findings indicate that sliding mode-based control is the most prevalent approach employed in the control of spherical robots.
The challenges associated with spherical robots encompass the limitations posed by their contact points, the integration of additional sensors, obstacle avoidance, controlled leaps, and control of non-holonomic systems. Addressing these challenges through targeted research efforts would drive advancements in spherical robotics, paving the way for more capable and versatile robotic systems in various applications.

This review emphasizes the importance of ongoing research and innovation in spherical robot design, locomotion, control, and embedded sensing. It emphasizes the potential of spherical robots in various applications and identifies areas for future exploration, such as refining control strategies, addressing challenges in obstacle avoidance, and comparing different approaches to modeling their dynamics.

Overall, the paper provides a comprehensive overview of spherical robot research, highlighting the current state of the field, key challenges, and potential avenues for further advancement.


\begin{acknowledgment}
This work was supported by NSERC Discovery Grant (RGPIN-2020-06121).
We would like to express our gratitude to Simon Bonnaud for his valuable support for the redaction of this manuscript.
\end{acknowledgment}

%
% ...

\bibliographystyle{asmems4}
\bibliography{main}
\end{document}